\documentclass[runningheads]{llncs}
\usepackage[T1]{fontenc}
\usepackage{graphicx}
\usepackage{booktabs}
\usepackage{multirow}
\usepackage{xurl}
\usepackage{comment}
\usepackage{subfig}
\usepackage[noadjust]{cite} 
\usepackage{xcolor} 
\usepackage{wrapfig}

\begin{document}


\title{
\makebox[\textwidth][c]{
VILAS: A VLA-Integrated Low-cost Architecture%
}\\[-0.15em]
with Soft Grasping for Robotic Manipulation
}

%
%
\author{Zijian An\inst{1} \and Hadi Khezam\inst{1} \and Bill Cai\inst{1} \and Ran Yang\inst{2} \and Shijie Geng\inst{3}\thanks{This work is independent of and outside of the author's work at Amazon.} \and Yiming Feng\inst{2} \and Yue Zheng\inst{1} \and Lifeng Zhou\inst{1}\thanks{Corresponding author}}
\authorrunning{Z. An et al.}
%
\institute{Drexel University, Philadelphia, PA 19104, USA \\
\email{\{za382,hk542,sc3568,yz894,lz457\}@drexel.edu}\and
Virginia Seafood Agricultural Research and Extension Center, Virginia Tech, 15 Rudd Ln, Hampton, VA 23669, USA\\
\email{\{ryang17,yimingfeng\}@vt.edu}\\
\and
Amazon Store Foundation AI (SFAI), 12 W 39th St, New York, NY 10018, USA\\
\email{\{jeykigung\}@gmail.com}}
%

\maketitle              
\begin{abstract}
We present VILAS, a fully low-cost, modular robotic manipulation platform 
designed to support end-to-end vision-language-action (VLA) policy learning 
and deployment on accessible hardware. The system integrates a Fairino FR5 
collaborative arm, a Jodell RG52-50 electric gripper, and a dual-camera 
perception module, unified through a ZMQ-based communication architecture 
that seamlessly coordinates teleoperation, data collection, and policy 
deployment within a single framework. To enable safe manipulation of fragile 
objects without relying on explicit force sensing, we design a kirigami-based 
soft compliant gripper extension that induces predictable deformation under 
compressive loading, providing gentle and repeatable contact with delicate 
targets. We deploy and evaluate three state-of-the-art VLA models on the VILAS 
platform: $\pi_0$, $\pi_{0.5}$, and GR00T N1.6. All models are fine-tuned 
from publicly released pretrained checkpoints using an identical demonstration 
dataset collected via our teleoperation pipeline.
Experiments on a grape grasping task validate the effectiveness of the proposed 
system, confirming that capable manipulation policies can be successfully 
trained and deployed on low-cost modular hardware. Our results further provide 
practical insights into the deployment characteristics of current VLA models 
in real-world settings.

\end{abstract}
\section{INTRODUCTION}
Robotic manipulation of fragile objects, such as fresh agricultural products, 
remains a challenging problem in both industrial and research settings 
\cite{ao2025end}. Unlike rigid industrial parts, fragile items such as tomatoes 
or berries possess low tolerance to excessive contact force and are easily damaged 
during grasping \cite{wen2020force, kultongkham2021design}. Reliable manipulation 
therefore requires careful coordination between perception, control, and mechanical 
design to ensure stable handling without inducing crushing or surface damage.

In addition to the intrinsic fragility of such objects, safe physical interaction 
often relies on precise regulation of contact forces. Conventional approaches 
achieve this through multi-axis force--torque sensors, tactile arrays, impedance 
control, or specialized end-effectors tailored for delicate handling 
\cite{li2023design, navas2024soft, ochoa2025design}. While effective, these 
sensing and control modules increase both hardware cost and system complexity, 
particularly when integrated into fully functional robotic learning platforms.

Soft and compliant grippers have emerged as a promising alternative, providing 
passive mechanical compliance that absorbs contact force variations without active 
sensing \cite{liu2023intelligent, shintake2018soft}. By offloading control 
complexity to the mechanical design itself, compliant end-effectors can enable 
gentle and repeatable grasping of delicate objects without relying on closed-loop 
force regulation \cite{doi:10.1126/scirobotics.abd6426}. However, existing soft 
gripper designs are often task-specific or require specialized fabrication 
processes, limiting their accessibility and integration into general-purpose 
low-cost manipulation platforms.

Recent advances in Vision-Language-Action (VLA) models have introduced a unified 
perception-to-action paradigm for robotic control \cite{kawaharazuka2025vision, 
brohan2023rt2}. By directly mapping multimodal observations to executable actions, 
VLA-based systems reduce reliance on manually engineered modular pipelines and 
enable flexible task generalization \cite{kim2024openvla, an2025claw, 
yang2025seqvla}. However, training and deploying such models typically depends on 
high-cost robotic platforms equipped with integrated hardware ecosystems 
\cite{zhao2023act, fu2024mobile, 
open_x_embodiment2023, octo2024}. As a result, the financial barrier for developing 
and experimenting with learning-based manipulation systems remains substantial 
\cite{mandlekar2018roboturk, khazatsky2024droid, dass2024telemoma}.

In this work, we present VILAS, a fully low-cost, modular robotic manipulation 
platform designed to address fragile object manipulation from a system-level 
perspective, with the explicit goal of enabling VLA-based learning on affordable 
hardware. Rather than relying on a unified proprietary robotic ecosystem, we 
construct a complete manipulation platform from independently selected, 
cost-efficient components integrated through a custom-developed control and data 
collection framework tailored for VLA training \cite{fang2024airexo}. To mitigate 
reliance on expensive force-sensing modules, we introduce a kirigami-based soft 
compliant extension attached to the end-effector, which passively absorbs contact 
force variations and provides intrinsic safety during grasping without additional 
sensing hardware. We further validate the platform by deploying and benchmarking 
three state-of-the-art VLA models on identical hardware and demonstration data, 
providing practical insights into their comparative deployment characteristics.

The primary contributions of this paper are as follows:
\begin{itemize}
\item We design and implement VILAS, a fully integrated low-cost robotic 
manipulation platform that supports the complete VLA workflow, covering 
teleoperation-based data collection, policy fine-tuning, and real-time 
deployment, using modular and affordable hardware components.
\item We design a mechanically compliant kirigami-based soft extension that 
facilitates safe manipulation of fragile objects without relying on explicit 
force sensing or complex force-control strategies.
\item We deploy and evaluate three state-of-the-art VLA models, namely $\pi_0$ 
\cite{black2410pi0}, $\pi_{0.5}$ \cite{pi05_2025}, and GR00T N1.6 
\cite{nvidia2025groot_n1, nvidia2025groot_n16}, on the VILAS platform using 
an identical demonstration dataset, and demonstrate that capable manipulation 
policies can be successfully trained and deployed on low-cost modular hardware, 
while providing empirical insights into the practical trade-offs across models.
\end{itemize}

\vspace{5pt}

\section{Related Work}
\noindent \textbf{Vision-Language-Action model.}
Recent advances in robotic learning have increasingly shifted from modular 
perception--planning--control pipelines toward unified learning-based frameworks 
\cite{brohan2023rt2, kawaharazuka2025vision}. 
VLA models represent a prominent direction within this paradigm, aiming to directly 
map visual observations and language instructions to executable robot actions through 
end-to-end policy learning \cite{kim2024openvla}. By jointly leveraging 
visual context and task-level language guidance, VLA models enable scalable and 
generalizable robotic control across diverse manipulation tasks. Existing approaches 
include transformer-based autoregressive action prediction \cite{zhao2023act} as well 
as diffusion or flow-matching formulations \cite{chi2023diffusion} for modeling 
continuous control outputs. Among these, $\pi_0$ combines a pre-trained 
vision-language model with a flow matching action expert to enable continuous action 
generation conditioned on visual observations and language instructions 
\cite{black2410pi0}, while $\pi_{0.5}$ extends this architecture with a heterogeneous 
mixture-of-experts design and large-scale internet data pre-training to improve 
generalization across diverse tasks and environments \cite{pi05_2025}. NVIDIA 
Isaac GR00T N1.6 similarly adopts a vision-language model backbone paired with a 
diffusion transformer head for action generation, and is further trained on thousands 
of hours of cross-embodiment teleoperation data to support generalization across 
heterogeneous robot platforms \cite{nvidia2025groot_n16}. In this work, we evaluate 
all three of these models deployed on a fully low-cost robotic manipulation platform, 
assessing their performance under end-to-end perception-to-action learning on modular 
hardware.

\vspace{15pt}

\noindent \textbf{Low-Cost Robotic Arm.}
A complete robotics learning system typically comprises several key components. 
First, the robot arm serves as the primary executor of manipulation tasks, 
providing the actuation and kinematic degrees of freedom required to interact 
with the environment \cite{fu2024mobile, zhao2023act}. Second, a teaching 
interface or teleoperation device is used for demonstration collection, allowing 
human operators to guide the robot's motion and generate high-quality ground-truth 
data for learning \cite{wu2024gello, fang2024airexo}. By physically controlling 
the teaching interface, the corresponding robot arm movements are recorded, forming 
the basis for imitation learning or other data-driven training paradigms. Third, 
one or more cameras form the visual perception module, capturing images or video 
streams that encode the scene context and object states observed during task 
execution \cite{wang2025robot, chi2023diffusion}. These visual inputs are essential 
for perception-driven control and policy learning.

There exist commercially available systems designed to support learning-based 
robotic manipulation research. A representative example is the ALOHA system 
\cite{zhao2023act}. ALOHA supports end-to-end imitation learning by collecting 
real-world demonstrations through a teleoperation interface and enables bimanual 
manipulation tasks on a mobile platform. The system integrates robotic hardware, 
teleoperation components, and perception modules into a unified setup tailored 
for data collection and policy training. Complete ALOHA hardware kits are 
available for purchase at approximately \$33,000 USD, providing a ready-to-deploy 
platform for real-world robotic learning experiments. However, ALOHA relies on 
Dynamixel servo motors, which are widely used in academic research platforms but 
are known to exhibit discrete stepping behavior and limited motion smoothness under 
continuous control. In contrast, the VILAS platform is built around the Fairino 
FR5, an industrial-grade collaborative arm driven by brushless DC motors, which 
produce inherently smoother and more continuous motion profiles. This mechanical 
characteristic is particularly beneficial for delicate manipulation tasks, where 
abrupt joint movements can compromise grasping precision or cause unintended 
contact forces on fragile objects.

\noindent \textbf{Soft Gripper}. 
There has been great development of compliant kirigami-based structures for soft robotic grippers, where strategically placed cuts allow thin sheets to undergo large, controlled shape transformations. In kirigami mechanisms, patterns such as parallel slits, staggered slit arrays, rotating square lattices, auxetic polygonal cuts, elliptical hole patterns, and periodic circular or create flexible ligaments that rotate and buckle under loading. This enables planar sheets to expand into three-dimensional grasping geometries. The mechanical response of these structures depends strongly on the geometry, spacing, and orientation of the cuts. Depending on these geometry cuts, this determines stiffness and deformation modes during stretching or bending \cite{article}. Several soft gripper designs exploit these mechanisms by transforming patterned sheets into curved gripping surfaces under simple actuation. Using a parallel slit kirigami pattern expands laterally to produce curvature that encloses objects, while hole pattern kirigami sheets generate shell like gripping structures through ligament rotation between perforations. A notable implementation is the kirigami shell gripper where a flat sheet with periodic hole arrays forms a three-dimensional shell when pulled by a single actuator, enabling adaptive grasping across objects of varying shapes and sizes \cite{doi:10.1126/scirobotics.abd6426}. Other studies have explored auxetic kirigami lattices and polygonal cut units to regulate surface curvature and improve contact area during grasping \cite{WU2025113410}. Kirigami grippers are typically fabricated using laser cutting of thin polymer sheets such as PET or polypropylene to produce precise slit or perforation geometries, recent work has demonstrated 3D printing approaches, including FDM and resin-based printing, to directly fabricate kirigami inspired compliant structures. In most implementations, the kirigami sheet is fixed to the actuation mechanism through mechanical clamping, adhesive bonding, or embedding within elastomer layers, allowing tendon driven tension or pneumatic inflation to stretch the patterned sheet and generate out of plane curvature buckling. These kirigami architectures enable lightweight, highly deformable grippers that conform to objects and distribute contact forces effectively, making kirigami a promising design strategy for adaptive soft robotic manipulation.


\section{System Overview}

\subsection{System Architecture}
\begin{figure}[!htbp]
    \centering
    \includegraphics[width=\textwidth]{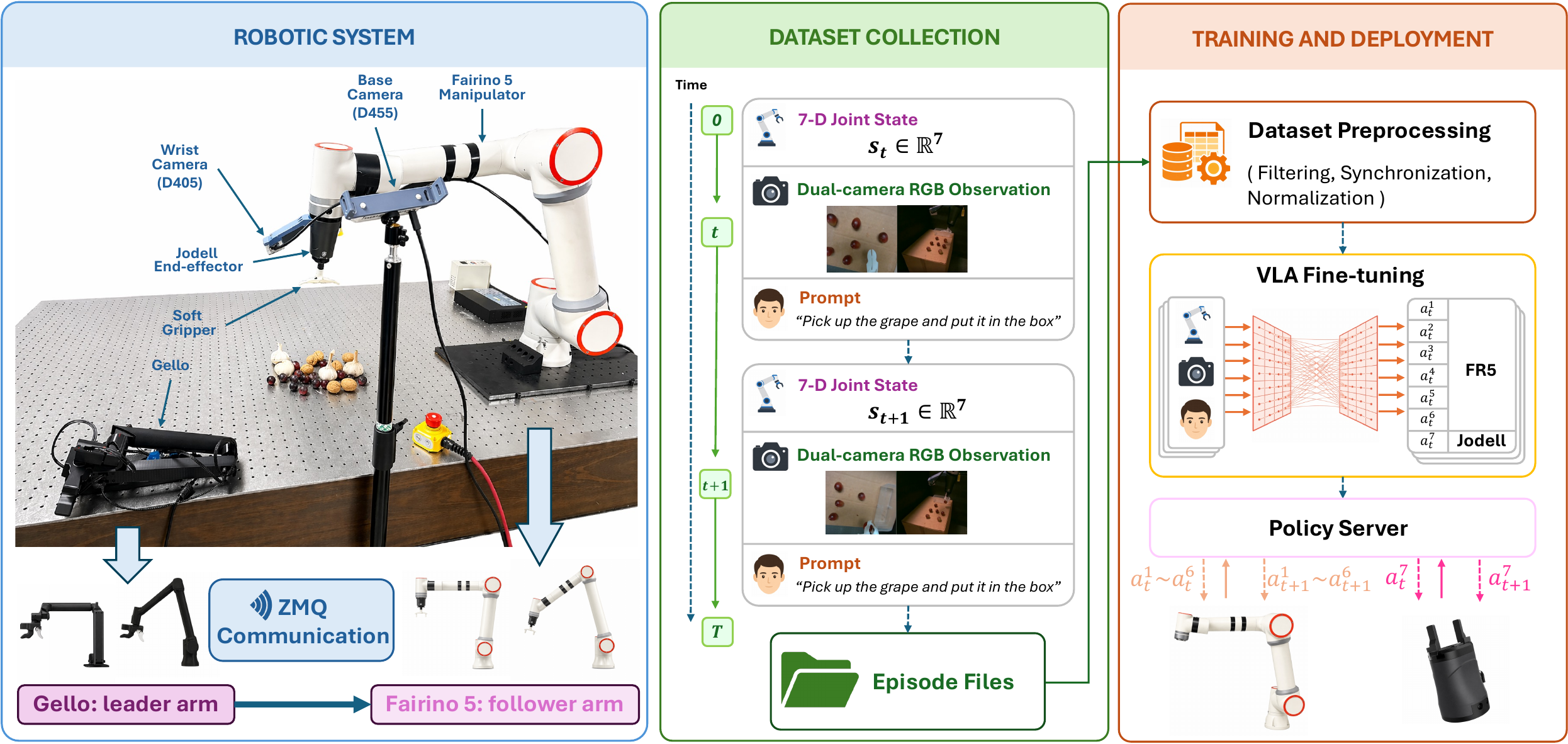}
    \caption{Overview of the VILAS system. (Left) The physical robotic platform. 
    (Center and Right) The data collection and training-and-deployment pipelines.}
    \label{fig:grasp_system}
\end{figure}

As illustrated in Figure~\ref{fig:grasp_system}, our VILAS system consists of 
a physical robotic platform and a two-stage operational pipeline covering data 
collection and policy deployment. The robotic platform centers on a 
Fairino FR5 manipulator equipped with a Jodell end-effector and a soft 
compliant gripper, observed by a base camera (RealSense D455) and a wrist 
camera (RealSense D405). During data collection, a human operator controls 
the FR5 via a GELLO leader arm through a ZMQ communication layer; at each 
timestep $t$, the system records a 7-dimensional joint state vector, 
dual-camera RGB observations, and a language prompt, which are stored as 
structured episode files. During deployment, the observation tuple consisting 
of joint states, dual-camera images, and a language prompt is passed to the 
policy model, which outputs an action chunk of shape $(50 \times 7)$ for 
$\pi_0$ and $\pi_{0.5}$, and $(16 \times 7)$ for GR00T N1.6, covering six 
arm joints and one gripper dimension, executed sequentially on the FR5 and 
Jodell gripper at 20\,Hz.

\vspace{5pt}

\vspace{-8pt}

\subsection{Hardware Platform}
The primary manipulator is a Fairino FR5, a six-degree-of-freedom industrial-grade 
collaborative robotic arm designed for real-world deployment in manufacturing and 
automation environments. With a 922\,mm reach, 5\,kg payload capacity, and 
$\pm$0.02\,mm pose repeatability (ISO 9283), the FR5 offers a level of mechanical 
precision and motion smoothness typically associated with professional-grade 
industrial systems. These specifications are well-suited for tabletop manipulation 
tasks such as pick-and-place and grasping, where moderate reach, sub-millimeter 
precision, and lightweight end-effector compatibility are the primary requirements. The FR5 communicates via standard TCP/IP and supports
ROS/ROS2 integration, enabling straightforward interfacing with learning-based
control pipelines. Critically, the FR5 achieves this capability at a hardware
cost of approximately \textbf{\$4{,}000\,USD}, placing it among the most cost-accessible
industrial cobots in its payload class.

 \vspace{3pt}

The end-effector is a Jodell RG52-50 electric parallel gripper, featuring an
adjustable stroke of 52\,mm, a controllable gripping force ranging from 2\,N to
50\,N, and a position repeatability of $\pm$0.02\,mm. The gripper is connected
to the control workstation via USB, using Modbus RTU as the underlying device
protocol, and supports independent control of position, force, and speed,
enabling compliant and precise interaction with objects of varying geometry and
fragility. Its IP65-rated enclosure and low operating noise ($<$40\,dB) further
make it suitable for close-range manipulation in unstructured environments.
At a hardware cost of approximately \textbf{\$1{,}500\,USD}, the RG52-50 provides
industrial-grade gripping performance at a fraction of the cost of comparable
force-controlled end-effectors.

\vspace{3pt}

For demonstration collection, we employ a GELLO teleoperation arm
\cite{wu2024gello}, a low-cost, open-source leader device constructed from
3D-printed parts and off-the-shelf Dynamixel servo motors, with a total
hardware cost of approximately \textbf{\$500\,USD}. GELLO is designed as a
kinematically equivalent replica of the target arm, such that each joint
of the leader maps directly and one-to-one to the corresponding joint of the
follower, eliminating the need for inverse kinematics and enabling intuitive,
joint-space teleoperation. Our GELLO configuration mirrors the
seven-degree-of-freedom structure of the FR5 system, comprising six arm joints
and one gripper joint, ensuring full kinematic correspondence between the leader
and follower during demonstration collection. The GELLO arm is connected to the
control workstation via USB, with joint angle readings transmitted using the
Dynamixel SDK.

\vspace{3pt}

The perception module consists of two Intel RealSense depth cameras selected to
provide complementary sensing coverage across the full manipulation workspace.
The RealSense D455 is mounted at a fixed overhead position to monitor the global
workspace. With an 86$^\circ$ field of view, a depth range of up to 4\,m, and a
depth error of less than 2\% at 4\,m, the D455 is well-suited for scene-level
object localization and workspace context. A RealSense D405 is mounted in close
proximity to the end-effector to capture fine-grained visual feedback during
gripper--object interaction. The D405 is optimized for short-range sensing,
operating at an ideal range of 7\,cm to 50\,cm with sub-millimeter depth
accuracy, and features a compact form factor of
42\,mm $\times$ 42\,mm $\times$ 23\,mm suitable for wrist-mounted deployment.
Both cameras are connected via USB and provide synchronized RGB and depth
streams through the Intel RealSense SDK 2.0. With a combined hardware cost
of approximately \textbf{\$1,000\,USD}, this dual-camera configuration
was selected as a cost-effective solution that satisfies the complementary
requirements of global scene awareness and local contact-level observation,
without relying on high-cost industrial vision systems.

\subsection{Soft Compliant End-Effector}
\begin{wrapfigure}{r}{0.5\textwidth}
    \vspace{-20pt}
    \centering
    \includegraphics[width=0.5\textwidth]{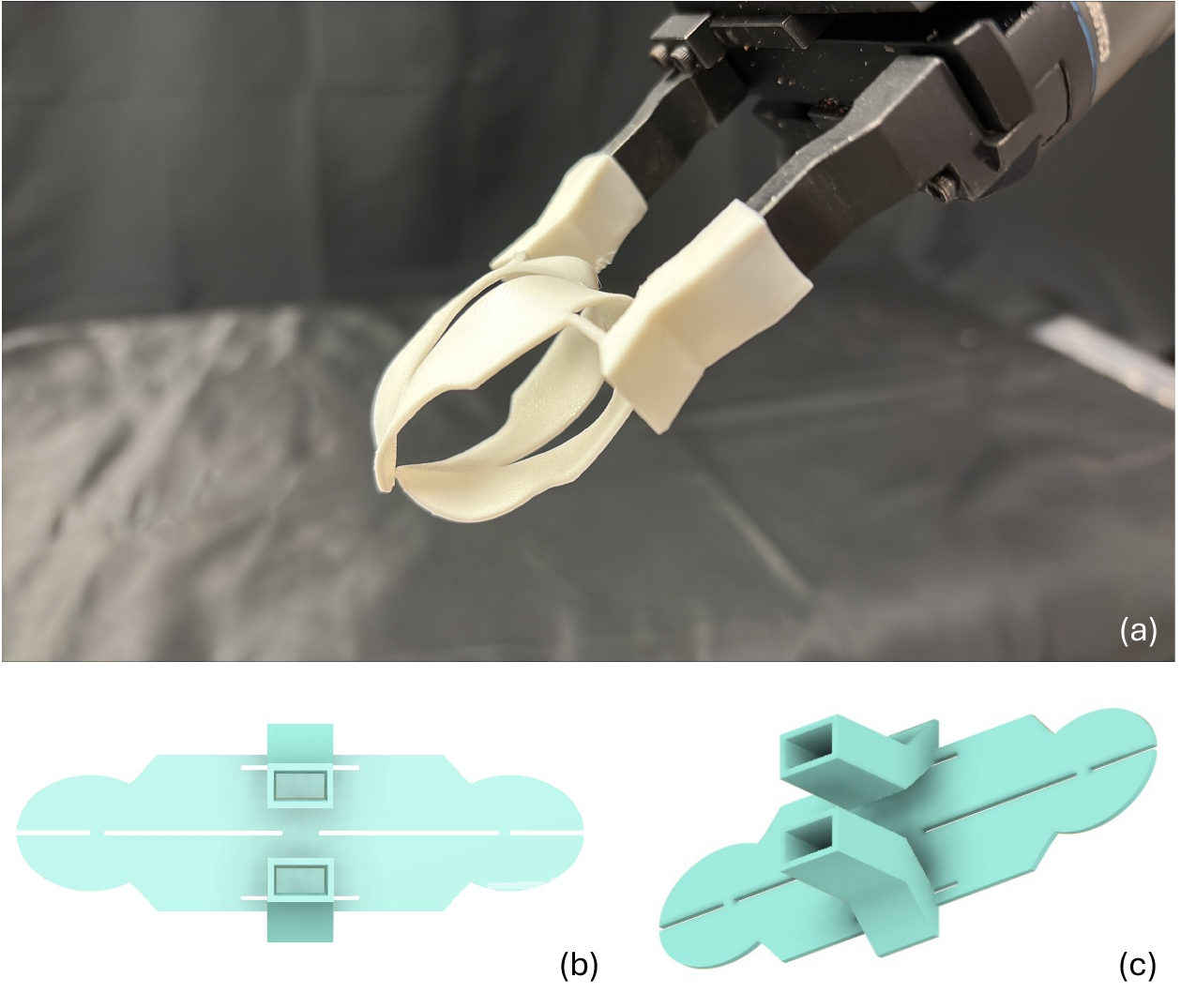}
    \caption{Kirigami structure design and experimental demonstration. 
    (a) Photograph of the fabricated kirigami structure buckled within 
    a gripper during testing, (b) Top view in Fusion 360, 
    (c) Isometric view in Fusion 360.}
    \label{softgripper}
    \vspace{-18pt}
\end{wrapfigure}
A low cost kirigami based pattern was developed to be used as a soft extension 
grabber for a safe and effective method of handling delicate objects as shown 
in Figure~\ref{softgripper}. Cut patterns were adapted from previously validated kirigami shell 
configurations \cite{doi:10.1126/scirobotics.abd6426}, with modifications to 
enable integration with a commercial parallel electric gripper (RG series). 
Expanding on previous work, the cuts on this kirigami structure side interfacing 
with the gripper were shortened to locally increase stiffness and reduce 
premature folding near the holder interface. The central cut lengths were 
extended to amplify deformation and promote smooth curvature formation under 
compressive loading. In addition, a kirgami extension was added at the distal 
end of the structure, increasing the overall length. This extension increased 
the effective contact region and enabled stable grasping of objects with varying 
diameters and curvatures as needed. The cut geometry was strategically designed 
to induce predictable deformation pathways, forcing the structure to bend in a 
prescribed direction under compressive input and improving repeatability across 
fabrication runs. Fabrication was performed using fused deposition modeling 
(FDM), printing the holder and kirigami structure as a single integrated PEBA 
component on a Bambu Lab H2D at 250 °C nozzle and 50 °C bed temperatures. PEBA 
was selected for its rapid "snap-back" elasticity, and the combined print 
reduced assembly and alignment errors, with each build taking ~45 minutes and 
70 g of filament to enable rapid, low-cost iteration.

The RG series gripper employs a two-finger parallel jaw configuration, which 
constrains the actuation to a single translational degree of freedom. To 
mechanically couple the kirigami structure to this gripper, a custom holder 
mechanism was integrated directly with the kirigami design, as demonstrated 
in Figure \ref{softgripper}. The holder attaches to the gripper fingers via 
a friction fit interface, enabling rapid mounting and removal without 
additional fasteners, adhesives, or permanent modifications. This design 
choice allows the kirigami gripper to be easily replaced or reconfigured for 
different experimental conditions. This inclined interface extends the 
effective closed length of the gripper which increases the achievable buckling 
curvature and expands the grasping envelope of the soft gripper, allowing it 
to envelop objects with larger diameters and irregular geometries while 
maintaining low contact forces. The inclined holder geometry also improves 
load distribution at the mounting interface, reducing stress concentration and 
improving durability over repeated actuation cycles. With this design we 
insured there is adequate buckling over long periods of actuation for a 
low-cost monolithic soft gripper. The kirigami extension is fabricated from 
off-the-shelf PEBA filament at a material cost of under \textbf{\$100\,USD} per unit, 
further reinforcing the low-cost design philosophy of the overall platform.

The total hardware cost of the VILAS platform, comprising the FR5 arm, 
Jodell RG52-50 gripper, GELLO teleoperation arm, dual RealSense cameras, 
and soft compliant extension, amounts to approximately \textbf{\$8{,}000\,USD}, 
which is substantially lower than the \$33{,}000\,USD cost of a comparable 
ALOHA system\footnote{To facilitate reproducibility and 
accessibility within the research community, we will release the full 
hardware specifications, software stack, and step-by-step assembly and 
deployment tutorials upon acceptance of this work.}.

\subsection{Communication and Control Architecture}

The FR5 manipulator communicates with the control workstation through an 
Ethernet-based TCP/IP interface provided by its controller, while a lightweight 
ZMQ wrapper standardizes access to low-level vendor commands and real-time 
state feedback by exposing unified arm command and observation endpoints to 
the teleoperation client. The Jodell gripper is similarly encapsulated within 
a ZMQ-based service layer, which abstracts the underlying device communication 
(connected via USB, using Modbus RTU protocol) and provides consistent control 
and state interfaces aligned with the arm module. 

\vspace{8pt}

\begin{figure}[!htbp]
    \centering
    \vspace{2pt}
    \includegraphics[width=0.8\textwidth]{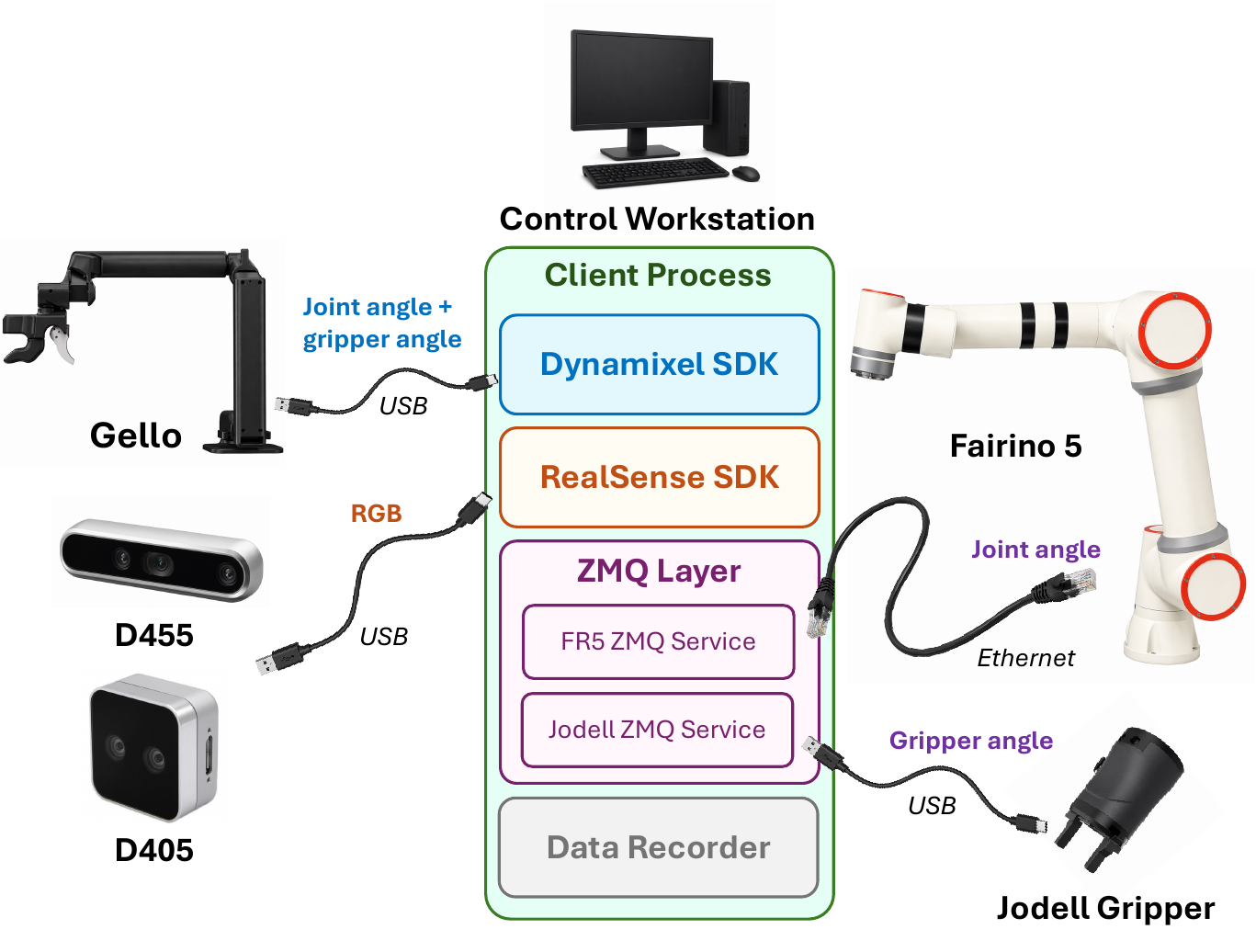}
    \vspace{-1pt}
    \caption{Communication architecture of the VILAS system during data collection. 
    The Control Workstation runs a unified Client Process comprising three layers: 
    the Dynamixel SDK for GELLO joint reading, the RealSense SDK for camera image 
    acquisition, and a ZMQ Layer that encapsulates communication with the FR5 arm 
    (via Ethernet) and Jodell gripper (via USB). A Data Recorder aggregates all 
    streams into synchronized episode files.}
    \label{fig:comm_architecture}
\end{figure}

The GELLO teleoperation arm 
operates within the client process and continuously supplies leader joint 
states; during teleoperation, these leader commands are transmitted through 
the ZMQ communication layer to the FR5 and Jodell modules, which execute the 
corresponding arm and gripper motions while returning synchronized feedback. 
This architecture establishes a bidirectional communication pipeline that 
coordinates leader input, follower execution, and state observation within a 
unified control framework. This architecture establishes a bidirectional communication pipeline that 
coordinates leader input, follower execution, and state observation within 
a unified control framework, as illustrated in Figure~\ref{fig:comm_architecture}.

\vspace{5pt}

The teleoperation loop operates at 83.3\,Hz, at which rate the GELLO arm 
reads operator joint angles via USB using the Dynamixel SDK and forwards 
them to the FR5 and Jodell modules through the ZMQ layer. Camera images are 
captured at 30\,Hz at a native resolution of 640$\times$480 and resized to 
224$\times$224 on the client side prior to transmission.

\vspace{5pt}

\vspace{-15pt}

During policy deployment, the control loop runs at 20\,Hz. Rather than 
invoking the model at every control step, we adopt an action chunking 
strategy in which a single inference call produces a chunk of 50 actions 
with shape $(50 \times 7)$, covering 6 arm joints and 1 gripper dimension. 
A broker module executes one action per control step from the cached chunk; 
once the chunk is exhausted after 2.5\,seconds, a new inference call is 
triggered. This decouples the high-frequency control loop from the relatively 
expensive model inference, which effectively runs at approximately 0.4\,Hz, 
and is consistent with the action chunking formulation introduced in $\pi_0$.

\vspace{5pt}

Model inference is handled by a dedicated GPU server connected to the robot 
workstation over a local network. The client is responsible only for 
observation collection and action execution, keeping the robot-side compute 
requirements minimal. For $\pi_0$ and $\pi_{0.5}$, inference is served via 
a WebSocket interface, with model selection controlled through a configuration 
parameter, allowing the two models to share an identical client-side interface. 
GR00T N1.6 is deployed via an independent ZMQ-based inference server, as its 
native deployment framework (NVIDIA Isaac GR00T) exposes a ZMQ endpoint by 
default, in contrast to the WebSocket-based serving interface provided by the 
$\pi_0$ codebase. While this results in heterogeneous communication protocols 
across models, the client-side observation format remains unified, and model 
switching requires only a change in the target server address and protocol 
adapter. In both cases, the observation passed to the server consists of a 
7-dimensional joint state vector (6 arm joint angles in radians plus a 
normalized gripper value) and two RGB image streams corresponding to the 
global workspace camera and the wrist-mounted camera, with a zero-filled 
placeholder retained for interface compatibility.

\subsection{Data Collection Pipeline}

During data collection, a human operator controls the FR5 manipulator in 
real time via the GELLO teleoperation arm. Joint angle commands from the 
GELLO leader arm are transmitted through the ZMQ communication layer to 
the FR5 and Jodell modules, which execute the corresponding motions while 
returning synchronized state feedback. At each time step, the 7-dimensional 
joint state vector and dual-camera RGB observations are recorded and stored 
as structured episode files. Following collection, the dataset is 
preprocessed and reformatted into the input schema required by each 
respective model's training framework.
\vspace{-4pt}

\section{Experiments}

\subsection{Experimental Setup}

For policy learning, we evaluate three VLA models: $\pi_0$, $\pi_{0.5}$, and 
GR00T N1.6. Rather than training from scratch, all models are initialized from 
their respective publicly released pretrained checkpoints and fine-tuned on the 
same demonstration dataset collected from our teleoperation pipeline, consisting 
of 100 demonstration episodes each comprising 1,200 frames. At each time step, 
the dataset includes synchronized visual observations from two cameras (a global 
workspace view and a close-range end-effector view), together with the 
corresponding robot joint states comprising six arm joints and one gripper joint 
(7-DoF in total). The recorded state--action pairs are formatted into structured 
datasets compatible with the respective training frameworks of each model. 
Fine-tuning for all three models is conducted on an NVIDIA H200 GPU for 50,000 
optimization iterations.

\subsection{Task Description}

We evaluate all models on a grape grasping task performed on a tabletop 
work\-space of approximately $30 \times 40$\, cm. At the beginning of each trial, 
10 grapes are placed at random positions within the workspace. The robot is 
tasked with sequentially grasping three grapes and depositing each into a 
small target box positioned at a fixed location on the table. No homing or 
reset motion is performed between consecutive grasps; upon completing one 
grasp-and-place cycle, the robot proceeds directly to the next target. The 
grape placement follows the same randomization protocol used during 
demonstration collection, ensuring consistency between training and evaluation 
conditions. To further assess cross-object generalization, we additionally evaluate all 
three models on a cherry grasping task in Section~\ref{sec:generalization}, 
using the same grape-trained policies without any additional fine-tuning.

\begin{figure}[!tp]
    \centering
    \includegraphics[width=\textwidth]{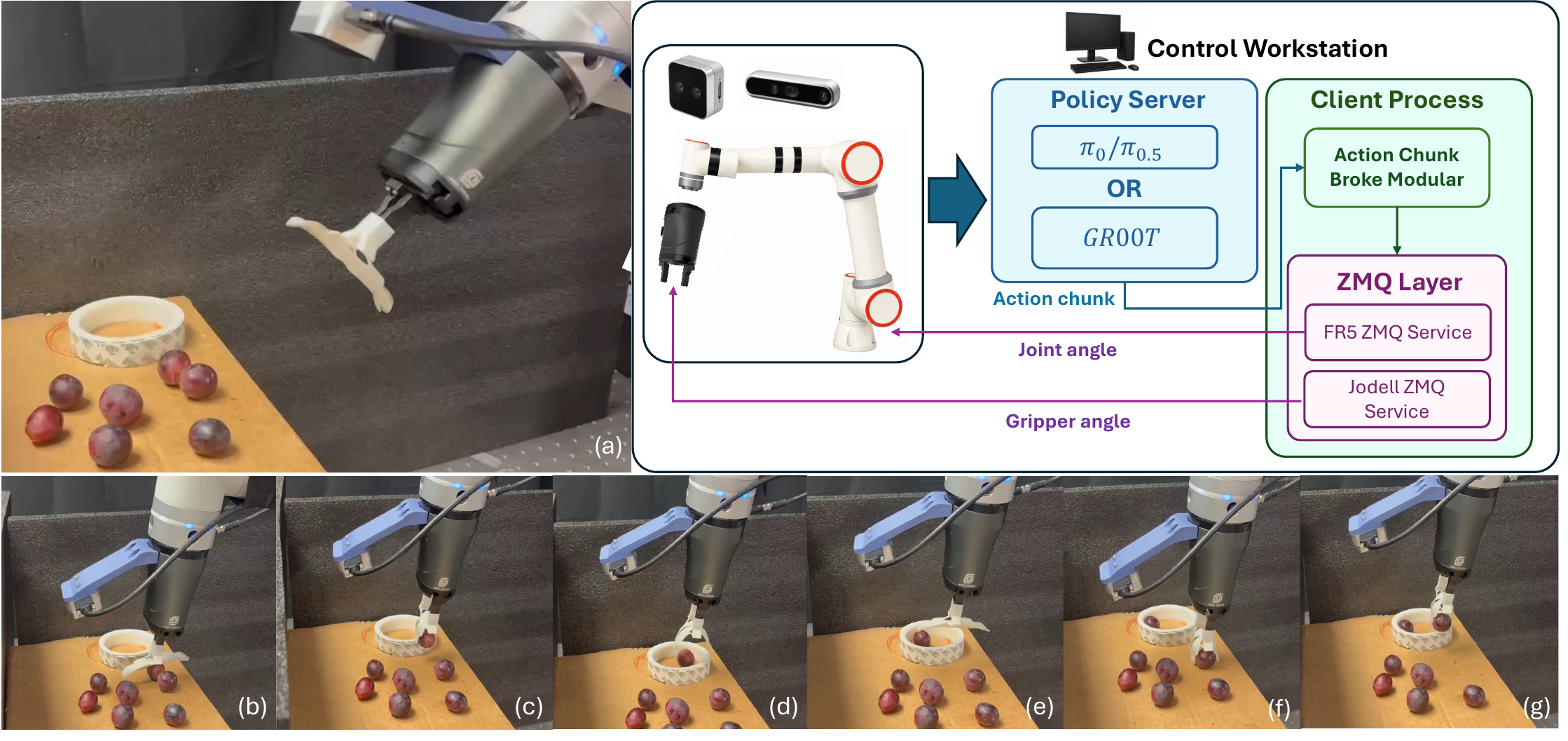}
    \caption{Policy deployment overview and representative execution sequence 
    using GR00T N1.6. (a) Initial system state, with the inset diagram 
    illustrating the deployment communication architecture: dual-camera 
    observations and robot joint states are forwarded to the Policy Server, 
    which outputs an action chunk dispatched via the ZMQ Layer to the FR5 arm 
    and Jodell gripper at 20\,Hz. (b)--(g) Sequential grasping execution: 
    (b) approaching the first target grape, (c) grasping the first grape, 
    (d) depositing it into the target box, (e) re-localizing the second target 
    grape, (f) grasping the second grape, and (g) completing the second deposit.}
    \label{fig:deployment}
\end{figure}

\subsection{Evaluation Protocol}

Each model is evaluated over 50 trials. We report two success metrics: 
\textit{single grasp success rate}, defined as the proportion of trials in 
which at least one grape is successfully grasped and deposited, and 
\textit{multi grasp success rate}, defined as the proportion of trials in 
which at least two consecutive grasps are completed successfully. These two 
metrics jointly capture both basic manipulation competence and the ability 
to sustain reliable performance across sequential grasping attempts.

\subsection{Results}

\begin{table}[t]
\centering
\caption{Comparison of inference latency and grasping success rate across 
three VLA models. All models are evaluated over 50 trials on the grape 
grasping task. Per-step cost is computed as mean latency divided by action 
horizon.}
\vspace{5pt}
\label{tab:model_comparison}
\begin{tabular}{lcccccccc}
\toprule
\multirow{2}{*}{Model} & 
\multicolumn{4}{c}{Inference Latency} & 
\multicolumn{1}{c}{Action} &
\multicolumn{1}{c}{Per-step} &
\multicolumn{2}{c}{Success Rate} \\
\cmidrule(lr){2-5} \cmidrule(lr){8-9}
& Mean & Median & Std & P95 & Horizon & Cost & 
Single (=1) & Multi ($\geq$2) \\
\midrule
$\pi_0$     & 73.8\,ms & 73.8\,ms & 0.4\,ms & 74.6\,ms & 50 & 
             1.48\,ms & 70\% & 22\% \\
$\pi_{0.5}$ & 82.8\,ms & 82.3\,ms & 2.4\,ms & 87.3\,ms & 50 & 
             1.66\,ms & 84\% & 36\% \\
GR00T N1.6  & 63.6\,ms & 63.7\,ms & 1.2\,ms & 65.0\,ms & 16 & 
             3.98\,ms & 82\% & 58\% \\
\bottomrule
\end{tabular}
\end{table}

Table~\ref{tab:model_comparison} summarizes the inference latency and grasping 
success rates of all three models. In terms of single grasp success, $\pi_{0.5}$ 
achieves the highest rate at 84\%, marginally outperforming GR00T N1.6 at 82\%, 
while $\pi_0$ lags behind at 70\%. However, the multi grasp metric reveals a 
more pronounced differentiation: GR00T N1.6 achieves a success rate of 58\%, 
substantially higher than $\pi_{0.5}$ at 36\% and $\pi_0$ at 22\%. This 
suggests that while $\pi_{0.5}$ and GR00T N1.6 exhibit comparable competence 
on individual grasps, GR00T N1.6 demonstrates superior consistency across 
sequential manipulation steps.

\vspace{5pt}

Figure~\ref{fig:deployment} presents the deployment communication architecture 
alongside a representative successful trial using GR00T N1.6. As shown in the 
inset of Figure~\ref{fig:deployment}(a), during deployment the robot receives 
no teleoperation input; instead, the policy server autonomously generates action 
chunks from visual observations and joint state feedback, which are executed 
at 20\,Hz via the ZMQ layer. The execution sequence in 
Figures~\ref{fig:deployment}(b)--(g) demonstrates the robot successfully 
completing two consecutive grasp-and-place cycles without intervention, 
consistent with GR00T N1.6's leading multi-grasp success rate of 58\%.

\vspace{5pt}

Regarding inference latency, GR00T N1.6 achieves the lowest mean latency at 
63.6\,ms, followed by $\pi_0$ at 73.8\,ms and $\pi_{0.5}$ at 82.8\,ms. All 
three models exhibit low standard deviations, indicating stable inference 
behavior suitable for real-time deployment. However, a direct latency 
comparison is complicated by differences in action horizon: $\pi_0$ and 
$\pi_{0.5}$ generate action chunks of 50 steps per inference call, while 
GR00T N1.6 produces chunks of 16 steps, a constraint imposed by its 
pretraining architecture. In our deployment, we adopt the maximum supported action horizon for each 
model: 50 steps for $\pi_0$ and $\pi_{0.5}$, and 16 steps for GR00T N1.6. Using the maximum chunk size minimizes inference 
frequency, reduces communication overhead between the robot workstation and 
the remote GPU server, and produces smoother motion trajectories by allowing 
the model to plan over a longer action horizon before re-querying. When 
normalized by action horizon, the per-step inference cost of GR00T N1.6 
(3.98\,ms) is notably higher than that of $\pi_0$ (1.48\,ms) and $\pi_{0.5}$ 
(1.66\,ms), suggesting that the lower raw latency of GR00T N1.6 is partly 
attributable to its shorter output sequence rather than a more efficient model 
architecture.

\subsection{Generalization to Other Objects}
\label{sec:generalization}
\begin{figure}[!tp]
    \centering
    \includegraphics[width=0.6\textwidth]{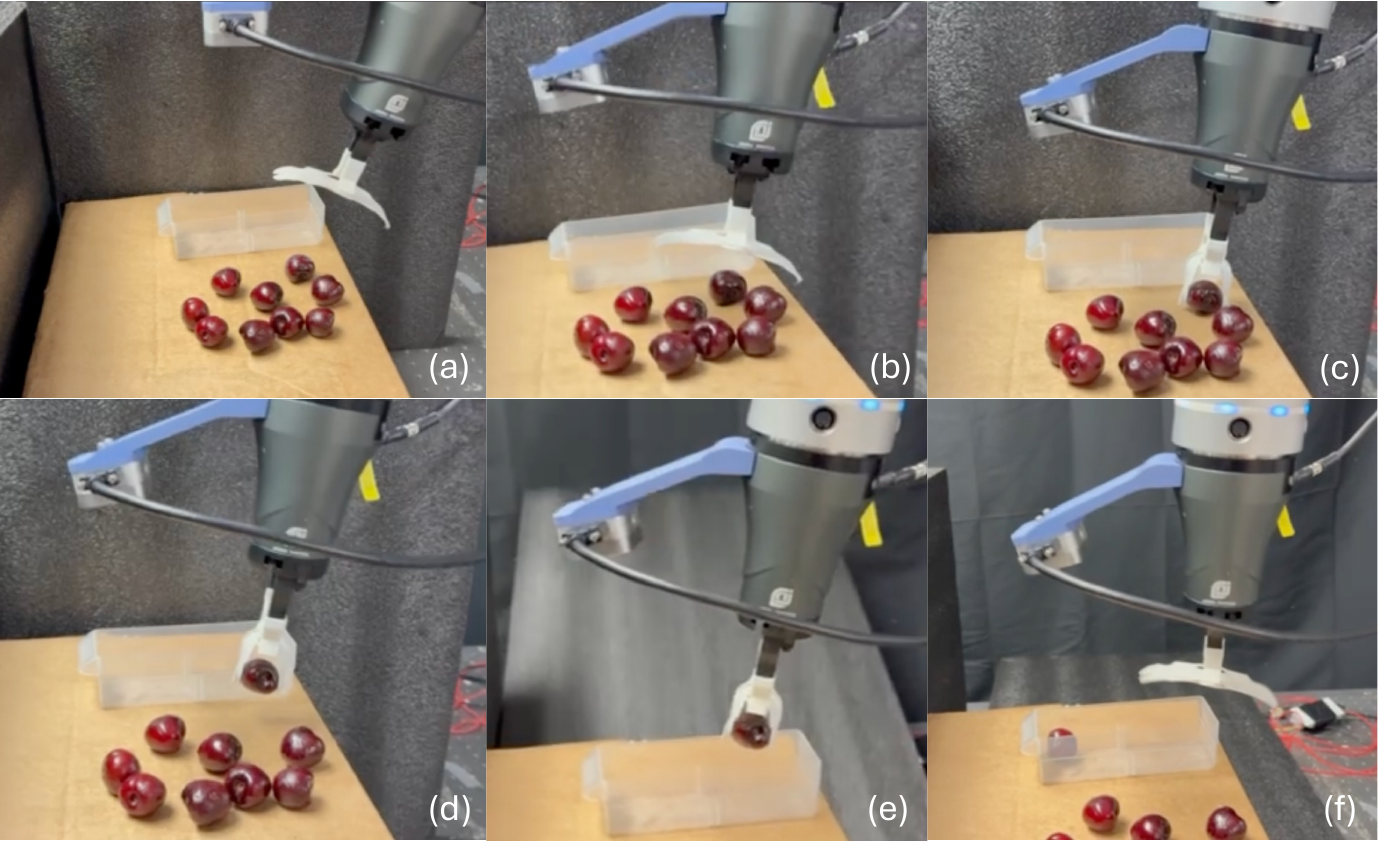}
    \caption{Execution sequence of the cherry grasping task using the $\pi_0$ 
    model without any modification to the model weights or fine-tuning data. 
    (a) Initial state with cherries randomly placed on the workspace. 
    (b) Approaching the target cherry. (c) Grasping the target cherry. (d) Transporting 
    the cherry toward the target box. (e) Depositing the cherry into the 
    target box. (f) Task completion.}
    \label{fig:cherry}
\end{figure}

To assess the generalizability of the learned policies beyond the training 
object, we evaluate all three models on a cherry grasping task using the 
same grape-trained policies without any additional fine-tuning. Only the 
language prompt and the physical object are replaced, with no modification 
to the model weights. Cherries present a distinct visual appearance from 
grapes, differing in size, color, and surface texture, making this a 
meaningful test of cross-object transfer. All three models demonstrate 
reasonable grasping performance on cherries, suggesting that the fine-tuned 
VLA policies capture manipulation primitives that generalize across visually 
distinct objects. Among the three, GR00T N1.6 shows comparatively more 
consistent performance, and its execution sequence is illustrated in 
Figure~\ref{fig:cherry} as a representative example.

\subsection{Failure Analysis}

We identify distinct failure modes for each model family based on qualitative 
observation across evaluation trials.


For $\pi_0$ and $\pi_{0.5}$, two primary failure patterns are observed. First, 
both models exhibit a tendency to return to the same spatial location after 
completing the first grasp, rather than searching for a new target. This 
suggests that the models fail to adequately update their spatial attention based 
on the current scene state, likely due to insufficient diversity in the 
demonstration data with respect to object layout variation. Second, gripper 
actuation consistency is poor in a subset of trials: the gripper intermittently 
releases during transport, causing the grasped object to drop, or repeatedly 
opens and closes at the target location without completing a successful grasp. 
This behavior indicates instability in the learned action chunks, particularly 
during the gripper closure phase.

\vspace{3pt}

GR00T N1.6 does not exhibit the same spatial repetition issue to the same 
degree, and gripper actuation is generally more consistent. However, it 
demonstrates a different failure pattern in multi-grasp trials: when searching 
for the second and third targets, the model tends to focus its search in the 
vicinity of the previous grasp location rather than exploring the broader 
workspace. As a result, multi-grasp success is sensitive to object placement 
density --- trials in which remaining grapes happen to be clustered near the 
prior grasp site benefit this behavior, while sparser layouts lead to missed 
targets. This spatial bias likely reflects a prior learned from the pretraining 
data distribution, where objects in sequential manipulation tasks tend to be 
spatially proximate.

\vspace{3pt}

These observations are consistent with the quantitative results in 
Table~\ref{tab:model_comparison}: the relatively small gap between single and 
multi grasp success rates for GR00T N1.6 (82\% vs.\ 58\%) compared to 
$\pi_0$ (70\% vs.\ 22\%) and $\pi_{0.5}$ (84\% vs.\ 36\%) suggests that 
GR00T N1.6 degrades more gracefully across sequential grasps, even if it 
is not immune to multi-target localization failures.
\vspace{-6pt}
\section{Conclusion and Future Work}
\vspace{-5pt}
In this work, we presented VILAS, a fully low-cost robotic manipulation platform 
designed to enable end-to-end VLA-based policy learning and deployment on 
modular, affordable hardware. The platform integrates a Fairino FR5 collaborative 
arm, a Jodell RG52-50 electric gripper, a custom kirigami-based soft compliant 
extension, and a dual-camera perception module, unified through a ZMQ-based 
communication architecture that supports seamless teleoperation, data collection, 
and policy deployment within a single framework. We deployed and benchmarked 
three state-of-the-art VLA models, namely $\pi_0$, $\pi_{0.5}$, and GR00T N1.6, 
all fine-tuned from publicly released pretrained checkpoints on an identical 
demonstration dataset. Experiments on a grape grasping task demonstrate that 
GR00T N1.6 achieves the highest multi-grasp success rate at 58\%, outperforming 
$\pi_{0.5}$ (36\%) and $\pi_0$ (22\%), while $\pi_{0.5}$ leads on single grasp 
success at 84\%. These results validate the effectiveness of VILAS as a 
reproducible and accessible testbed for VLA evaluation on real hardware.

Several directions remain for future work. On the task side, we plan to extend 
evaluation to more complex manipulation scenarios requiring longer-horizon 
reasoning and greater environmental diversity. On the hardware side, future 
iterations will explore soft gripper designs with improved lateral stiffness 
to enable reliable grasping in cluttered scenes.
\section{Acknowledgement}

The authors gratefully acknowledge Drexel Longsview Award for their generous support.

\newpage
%
%
%
\bibliographystyle{splncs04}
\bibliography{refs}

\end{document}